\pdfoutput=1

\documentclass[11pt]{article}

\usepackage[]{emnlp2021}
\usepackage{multirow}
\usepackage{hyperref}

\usepackage{times}
\usepackage{latexsym}
\usepackage{amsmath}
\usepackage{algorithm}
\usepackage{algpseudocode}
\usepackage{graphicx}
\usepackage{subfig}

\usepackage[T1]{fontenc}

\usepackage[utf8]{inputenc}

\usepackage{microtype}

\title{A Split-then-Join Approach to Abstractive Summarization for Very Long Documents in a Low Resource Setting}

\author{Lhuqita Fazry \\
Department of Computer Science \\
Universitas Indonesia \\
Depok 16424, Indonesia \\
\texttt{lhuqita.fazry@ui.ac.id}}

\begin{document}
\maketitle
\begin{abstract}
    \texttt{BIGBIRD-PEGASUS} model achieves \emph{state-of-the-art} on abstractive text summarization for long documents. However it's capacity still limited to maximum of $4,096$ tokens, thus caused performance degradation on summarization for very long documents. Common method to deal with the issue is to truncate the documents. In this reasearch, we'll use different approach. We'll use the pretrained \texttt{BIGBIRD-PEGASUS} model by fine tuned the model on other domain dataset. First, we filter out all documents which length less than $20,000$ tokens to focus on very long documents. To prevent domain shifting problem and overfitting on transfer learning due to small dataset, we augment the dataset by splitting document-summary training pair into parts, to fit the document into $4,096$ tokens \footnote{Source code available on \href{https://github.com/lhfazry/SPIN-summ}{https://github.com/lhfazry/SPIN-summ}}.
\end{abstract}

\section{Introduction}

Text summarization is a downstream and challenging task in natural language processing. The goal is to produce the shorter version of a document while preserving its information content \cite{nenkova_automatic_2011}. Producing the shorter version is not straightforward. A model needs to understand whole document then generate some text that can summarize the document well.

Currently, there are two common approaches to text summarization: \emph{extractive} and \emph{abstractive} \cite{see_get_2017}. From both approaches, extractive summarization is the simplest. The summary is produced by extracting salient sentences from the original text \cite{huang_efficient_2021}. In contrast, abstractive summarization attempts to produce a bottom-up summary \cite{rush_neural_2015}, thus yield some novel texts on the generated summary.

The majority of past work has been focused on extractive text summarization \cite{kupiec_trainable_1995,poibeau_automatic_2013,klymenko_automatic_2020}. This is caused by the difficulty of text generation task which is the backbone of abstractive summarization. \emph{Sequence to sequence} model introduced by \cite{sutskever_sequence_2014} solved the text generation problem by using a multilayered Long Short-Term Memory (\texttt{LSTM}) to map the input sequence
into a vector of a fixed dimensionality. The success of the model leading to the development of derivative models in abstractive summarization \cite{chopra_abstractive_2016,nallapati_abstractive_2016,rush_neural_2015,zeng_efficient_2016}.

Recently, \emph{transformers} \cite{vaswani_attention_2017} outperforms RNN-based model. The transformers architecture enable parallelization in sequences processing, so it can utilize the full power of recent advance in hardware accelerator like \texttt{GPU} and \texttt{TPU}. \texttt{BERT}, then popularized the use of transformers-based pretrained language model \cite{devlin_bert_2019}. Further improvement on \texttt{BERT}, yields to \emph{state-of-the-art} transformers-based model on text summarization like \texttt{BART} \cite{lewis_bart_2019} and \texttt{PEGASUS} \cite{zhang_pegasus_2020}.

However, due to the mechanism of \emph{self attention}, the memory and computational requirements of transformers-based models grow quadratically with sequence length \cite{child_generating_2019}. This limitation prevents such models to be used in summarization on long documents.

\section{Literature Review}

Formally, the problem of abstractive text summarization can be formulated as follow. Given an input consists of $a$ pairs of documents and summaries:  $\{(X_1,Y_1),(X_2,Y_2),\dots,(X_a,Y_a)\}$. Each documents and summaries are then tokenized into $X_i=(x_i,\dots,x_m)$ and $Y_i=(Y_1,\dots,y_n)$ respectivelly. Therefore, the model needs to generate an abstractive summary of $\hat{Y}=(\hat{y}_1,\dots,\hat{y}_{n'})$ \cite{zhu_enhancing_2021}.

One of the most common used pretrained model on abstractive text summarization among nlp researcher is \texttt{BART} \cite{lewis_bart_2019}. \texttt{BART} can be seen as combination between Bidirectional encoder (\texttt{BERT}) and Auto-Regresive decoder (\texttt{GPT}) \cite{radford_improving_nodate}. This combination yields an effective text generation model. \texttt{BART} then can be fine tuned to solve board range of abstractive task including dialogue, question-answering and summarization. The fine tuned version achieve \texttt{SOTA} on specific task. One of the most factor behind the effectiveness of \texttt{BART} is the pretaining stages. \texttt{BART} is pretrained on \texttt{GLUE} \cite{wang_glue_2019} and \texttt{SQuAD} \cite{rajpurkar_squad_2016} datasets within two training stages (i) text is corrupted with an arbitrary noising function, and (ii) model is trained to reconstruct the original text.

Rely on \texttt{BERT} as the encoder, \texttt{BART} inherits maximum input sequence length problem. Maximum tokens that can be processed by \texttt{BART} is $1,022$ tokens \cite{gaskell_summarization_2020}. This condition making \texttt{BART} not suited for long documents summarization. The simplest and common solution is to truncate the input document and only use first $1,022$ tokens. This solution can result bad performance on some documents because the important contents may spread across position. 

Another solution is using Divide-ANd-ConquER (\texttt{DANCER}) methods \cite{gidiotis_divide-and-conquer_2020}. \texttt{DANCER} split the documents into parts which length are sufficient to be feed into the model. The method then automatically split the respective target summary into parts and pairs each of these parts to the appropriate part of the document, in order to create distinct target summaries for each document part. Splitting the document is easy. In constrast, splitting the target summary is not straightforward.

Inspired by \texttt{BART}, \cite{zhang_pegasus_2020} studied another approach on pre-training objective, yielding \texttt{PEGASUS} model that achieve \emph{state-of-the-art} on abstractive text summarization. \texttt{PEGASUS} propose novel pre-training objective called \emph{Gap Sentences Generation} (GSG). \texttt{PEGASUS} is trained by selecting and masking $m$ sentences from documents, then the model learn to predict those sentences using the remaining unmasked sentences. There are 3 sentence selection strategies proposed by the author: i) Randomly select $m$ sentences, ii) Select $m$ first sentences, iii) Compute \texttt{ROUGE1-F1} score for all sentences then select top-$m$ scored sentences. 

There are also some extensive researches on efficient attention to improve transformers capability on longer sequences. One of the promising model is \texttt{BIGBIRD} \cite{zaheer_big_2021}. \texttt{BIGBIRD} replace full attention mechanism on transformers by using sparse attention. Sparse attention in \texttt{BIGBIRD} is a combination of random attention, sliding window attention and global attention. By using graph theory, the author shows that such combination approximate performance of full attention. Thus, \texttt{BIGBIRD} reduce computation complexity from $O(n^2)$ to $O(n)$ while preserving the full attention performance. The author then apply \texttt{BIGBIRD} architecture to \texttt{PEGASUS} model. This approach yielding \texttt{BIGBIRD-PEGASUS} model which achieve state-of-the-art on text abstractive summarization task. It achieves \texttt{ROUGE-1} score $46.63$, $46.32$, $60.64$ for \emph{arXiv} \cite{cohan_discourse-aware_2018}, \emph{PubMed} \cite{cohan_discourse-aware_2018} and \emph{BigPatent} \cite{sharma_bigpatent_2019} datasets respectivelly.

Although \texttt{BIGBIRD} successfully increase maximum input tokens on transformers, it's still limited to $4,096$ tokens \cite{zaheer_big_2021}. Some datasets like arXiv and PubMed has an average $5,179$ token length, while BigPatent has $3,629$. This is the reason why \texttt{R1} score of \texttt{BIGBIRD-PEGASUS} on BigPatent is better than on arXiv and PubMed. 

\section{Methodology}

The mode common method to deal with long documents is truncate the documents. In this reasearch, we'll use different approach. We'll evaluate the pretrained \texttt{BIGBIRD-PEGASUS} model on very long documents (more than $20,000$ tokens). We used subset of arXiv and BigPatent datasets that has document length more than $20,000$ tokens. We download the pretained model of \texttt{BIGBIRD-PEGASUS} from HuggingFace \cite{wolf_huggingfaces_2020} and fine tuned it.

In order to give \texttt{BIGBIRD-PEGASUS} model the ability to capture very long documents, we split each documents and target summaries into parts in the training steps, inspired by the \texttt{DANCER} method \cite{gidiotis_divide-and-conquer_2020}. Different with the \texttt{DANCER} method, we split the summaries into same number part of documents, instead of splitting by sentence. Then we match each document part with summary part with the highest \texttt{ROUGE-L}. We named our proposed approach to SPlit-then-joIN (\texttt{SPIN}). Our approach consists of two steps.

\subsection{Training Step (Split)}

As stated before, we split each documents and target summaries into parts in the training steps. In detail, each document parts can have length maximum of $4,096$ tokens. Assume we have document-summary training pair which document length is $K$ tokens, where $K$ greater than $4,096$. The the document is splitted into $L$ parts $(p^1,\dots,p^L)$ where $L=\lceil\frac{K}{4,096}\rceil$. Each part $p^l$ can have length maximum of $4,096$ tokens and can be represented as a list of M sentences $p^l=(p_1^l,\dots,p_M^l)$. We created 3 \texttt{SPIN} variants regarding to the target summary splitting.

\subsubsection{\texttt{SPIN} 1}
In \texttt{SPIN} 1, the target summary is also splitted into $L$ parts $(s_1,\dots,s_L)$, same with the document. Given any two sequences of words $x=(x_1,\dots\,x_P)$ and $y=(y_1,\dots,y_P)$ with lengths $P$ and $Q$ respectivelly, we define $LCS(x,y)$ as the longest common sub-sequence of $x$ and $y$. Then \texttt{ROUGE-L} recall between $x$ and $y$ can be computed as follows:
\begin{align}
    R_{LCS}(x,y) &= \frac{LCS(x,y)}{Q} \label{eq:recall_lcs}
\end{align}

Using formula \ref{eq:recall_lcs}, then we compute $R_{LCS}$ between each parts of document $p^l$ and each parts of summary $s_n$. For any document part $p^l$, we pair it with summary $s_n$ that has greatest value of $R_{LCS}(p^l,s_n)$. We repeat this step untill all document parts have been paired with one summary part. Note that, the pairing is one-to-one because document and summary has same number of parts. This splitting process then repeated untill all documents in the training set has been paired. This process is summarized in Algorithm \ref{alg:1}.

\subsubsection{\texttt{SPIN} 2 \& \texttt{SPIN} 3}
\begin{figure*}[t]
    \centering
    \begin{tabular}{cc}
      \includegraphics[height=7cm,keepaspectratio]{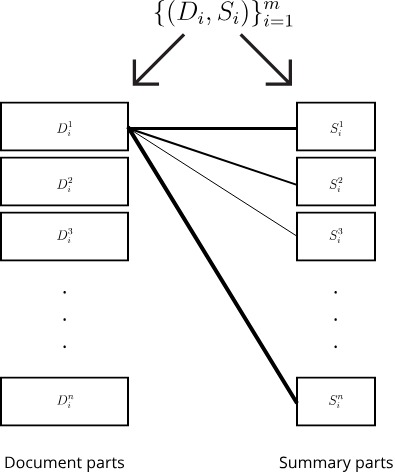} &   \includegraphics[height=7cm,keepaspectratio]{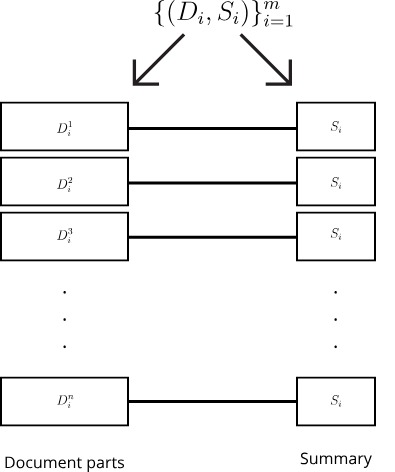} \\
    (a) \texttt{SPIN} 1 & (b) \texttt{SPIN} 2 \& 3 
    \end{tabular}
    \caption{Document-summary pairing strategy between \texttt{SPIN} 1 and \texttt{SPIN} 2 \& 3}
    \label{fig:1}
\end{figure*}

Unlike \texttt{SPIN} 1, the target summary is not splitted in in \texttt{SPIN} 2 and \texttt{SPIN} 3. Instead, we paired each document parts into the full target summary. Figure \ref{fig:1} ilustrates both document-summary pairing strategies.

\begin{algorithm}
    \caption{\texttt{SPIN} algorithm}\label{alg:1}
    \begin{algorithmic}[1]
    \Require $\mathbf{A} = \{(D_i,S_i)\}^m_{i=1}$, document - summary pairs
    \Ensure $\hat{\mathbf{A}} = \{(\hat{D}_i, \hat{S}_i)\}^n_{i=1}$, augmented document - summary pairs ($n > m$)
    \State $\hat{\mathbf{A}} \gets \{\}$
    
    \For{$i = 1, \cdots, m$}
        \State $l_k \gets len(D_i.split())$
        \State $l_s \gets len(S_i.split())$
        \State $n\_part \gets \lceil\frac{K}{4096}\rceil$

        \If{$K > 4096$}
            \State $J \gets range(0, l_k, 4096)$
            \State $\mathbf{B} \gets \{D_i(j:j+4096)\: \mathbf{for}\: j\in J \}$

            \State $step \gets \frac{l_s}{n_parts}$

            \State $J \gets range(0, step \times n\_parts, step)$
            \State $\mathbf{T} \gets \{S_i(j:j+step)\: \mathbf{for}\: j\in J \}$

            \For{$b \in \mathbf{B}$}
                \State $s \gets \underset{t \in \mathbf{T}}{argmax}\: R_{LCS} (b,t)$
                \State $\hat{\mathbf{A}}.append((b,s))$
                \State $\mathbf{T}.remove(s)$
            \EndFor
        \EndIf
    \EndFor
    \end{algorithmic}
\end{algorithm}

\subsection{Generating Step (Join)}

After the model has been trained, it's now can be used for generating summary to any unseen documents. The generating step is as follow. Given document-summary pair $(D, S)$ where length of $D$ is greater than $4,096$. Following the training step, we split $D$ into parts $\{\hat{D}_i\}^m_{i=1}$ where $\hat{D}_i$ can have maximum length of $4,096$. Then we use the trained model to generating summary for every part $\hat{D}_i$. 

In \texttt{SPIN} 1 and \texttt{SPIN} 2, the generated summaries $\{\hat{S}_i\}^m_{i=1}$ are then concatenated, resulting a single summary $\hat{S}$. The proses of generating summary $\hat{S_i}$ are independence each others, so it can be done in parallel. To evaluate the model performance, we calculated \texttt{ROUGE} metric between $S$ and $\hat{S}$.

In \texttt{SPIN} 3, we calculated \texttt{ROUGE-L} score between $\{\hat{D}_i\}^m_{i=1}$ and their respective generated summary $\{\hat{S}_i\}^m_{i=1}$. Then we choose $\hat{S}_i$ that has greatest score as final generated summary $\hat{S}$.

Filtering the datasets to only include documents which length more than $20,000$ tokens making the training data size reduced drastically. Small dataset lead to domain shifting problem, and overfitting on transfer learning \cite{chen_meta-transfer_2021}. Our method effectively handle this situation because it can augmented the training sample by splitting very long document yielding more sample data for the model.

\section{Experiment}

We've conducted some extensive experiments to evaluate our proposed method. We used 1 Core GPU of Tesla V100 for training and inferencing the model. We also trained the baseline model in the same environment to bring a fair evaluation. In this section we'll explain the datasets and the models in detail.

\subsection{Data}

We used two large-scale and commonly used datasets for summarization task, arXiv \cite{cohan_discourse-aware_2018} and BigPatent \cite{sharma_bigpatent_2019} datasets. Both are publicly available. We used these datasets because they aligned with our research focus on very long documents. We've conducted an extensive reasearch on summarization datasets and found that only arXiv anda BigPatent that contain enough samples which have more than $20,000$ token length.

The arXiv dataset contains scientific papers and their abstract that taken directly from arXiv repository. The BigPatent dataset contains patent documents in the United States. Table \ref{table:datasets} summarizes the datasets that we used in this research. We filter out all documents which length less than $20,000$ to focus on very long documents.

\begin{table}
    \centering
    \begin{tabular}{cccc}
    \hline
    \textbf{Dataset} & \textbf{Parameter} & $\mathbf{|D|}$ & $\mathbf{|S|}$ \\
    \hline
    \multirow{5}{*}{arXiv} & count & $3,533$ & $3,533$\\
    & mean & $27,588.19$ & $308.37$\\
    & 50\% & $24,252$ & $268$ \\
    & 75\% & $29,236$ & $358$ \\
    & max & $299,324$ & $11,658$\\
    \hline
    \multirow{5}{*}{BigPatent} & count & $6,589$ & $6,589$\\
    & mean & $28,868.80$ & $133.48$\\
    & 50\% & $25,488$ & $121$ \\
    & 75\% & $31,570$ & $161$ \\
    & max & $143,053$ & $4,063$\\
    \end{tabular}
    \caption{Statistics of subset on arXiv \cite{cohan_discourse-aware_2018} and BigPatent \cite{sharma_bigpatent_2019} documents which token length more than $20,000$ token. $|D|$ and $|S|$ refer to document and summary tokens length respectivelly}
    \label{table:datasets}
\end{table}

Following the proposed method, we then augmented the data by splitting the document-summary into parts. Table \ref{table:datasets2} summarizes the datasets after we splitted it.

\begin{table}
    \centering
    \begin{tabular}{cccc}
    \hline
    \textbf{Dataset} & \textbf{Parameter} & $\mathbf{|D|}$ & $\mathbf{|S|}$ \\
    \hline
    \multirow{5}{*}{arXiv} & count & $24,481$ & $24,481$\\
    & mean & $3,808.09$ & $40.73$\\
    & 50\% & $4,096$ & $34$ \\
    & 75\% & $4,096$ & $50$ \\
    & max & $4,096$ & $1,168$\\
    \hline
    \multirow{5}{*}{BigPatent} & count & $49,828$ & $49,828$\\
    & mean & $3,817.46$ & $17.65$\\
    & 50\% & $4,096$ & $15$ \\
    & 75\% & $4,096$ & $22$ \\
    & max & $4,096$ & $313$\\
    \end{tabular}
    \caption{Statistics of augmented dataset version. $|D|$ and $|S|$ refer to document and summary tokens length respectivelly}
    \label{table:datasets2}
\end{table}

\subsection{Model}

As stated before, we used \texttt{BIGBIRD-PEGASUS} model architecture and pre-trained weight from HuggingFace \cite{wolf_huggingfaces_2020}. This pre-trained model is trained on summarization task for PubMed \cite{cohan_discourse-aware_2018} dataset, so both datasets (arXiv and BigPatent) are unseen data for this model. \texttt{BIGBIRD-PEGASUS} is a large model with $576,891,904$ parameters in total. 

We set the \emph{embedding\_size} for the model to $4,096$ so it can process input up to $4,096$ tokens. To fit this model on single GPU Tesla V100 when training, we can only use batch size of $4$. We can increase the batch size up to $64$ but we have to activate \emph{gradient\_checkpoint}. Activate it will slow down the training time to $20\%$. With these setting, we get an average $50$s per iteration.

We've trained \texttt{BIGBIRD-PEGASUS} using our method on arXiv and BigPatent datasets. We'll compare the result using standard training. In the standard training scenario, we truncated all documents into $4,096$ token length. 

\section{Result \& Discussion}
\begin{table*} [t]
    \centering
    \begin{tabular}{l|ccc|ccc}
    \hline
    \multirow{2}{*}{\textbf{Method}} & \multicolumn{3}{c|}{\textbf{arXiv}} & \multicolumn{3}{c}{\textbf{BigPatent}} \\
    & R1 & R2 & RL & R1 & R2 & RL \\
    \hline
    BigBird-Pegasus & $39.6$ & $13.2$ & $21.9$ & $23$ & $6.7$ & $17.2$ \\
    SPIN 1 & $11.7$ & $1.5$ & $8.9$ & $9.9$ & $1.9$ & $9.1$ \\
    SPIN 2 & $21.2$ & $8.7$ & $12.1$ & $12.9$ & $4.9$ & $9.4$ \\
    SPIN 3 & $\mathbf{41.7}$ & $\mathbf{15.6}$ & $\mathbf{26.0}$ & $\mathbf{35.6}$ & $\mathbf{15.2}$ & $\mathbf{28.0}$ \\
    \hline
    \end{tabular}
    \caption{Result comparison on dataset BigPatent. R1, R2 and RL refer to $\mathbf{ROUGE-1}$, $\mathbf{ROUGE-2}$ and $\mathbf{ROUGE-L}$ metrics respectivelly}
    \label{table:result}
\end{table*}

Table \ref{table:result} summarize the result of the experiments for arXiv and BigPatent datasets respectivelly. The result show that \texttt{SPIN} 3 is better than \texttt{SPIN} 1, \texttt{SPIN} 2 and \texttt{BIGBIRD-PEGASUS}. This result shows that the most important information not always contained in the head of document. Actually it can be in any parts, may be in the middle or in the end of the document. 

The reason why \texttt{SPIN} 2 is not perform well is because by splitting summaries with a fixed length equal to the number of document parts make the document-summary pair become unrelevant so the score is decreased.

\bibliography{anthology,custom,PIL}

\begin{thebibliography}{27}
\expandafter\ifx\csname natexlab\endcsname\relax\def\natexlab#1{#1}\fi

\bibitem[{Chen and Shuai(2021)}]{chen_meta-transfer_2021}
Yi-Syuan Chen and Hong-Han Shuai. 2021.
\newblock \href {http://arxiv.org/abs/2102.09397} {Meta-{Transfer} {Learning}
  for {Low}-{Resource} {Abstractive} {Summarization}}.
\newblock \emph{arXiv:2102.09397 [cs]}.
\newblock ArXiv: 2102.09397.

\bibitem[{Child et~al.(2019)Child, Gray, Radford, and
  Sutskever}]{child_generating_2019}
Rewon Child, Scott Gray, Alec Radford, and Ilya Sutskever. 2019.
\newblock \href {http://arxiv.org/abs/1904.10509} {Generating {Long}
  {Sequences} with {Sparse} {Transformers}}.
\newblock \emph{arXiv:1904.10509 [cs, stat]}.
\newblock ArXiv: 1904.10509.

\bibitem[{Chopra et~al.(2016)Chopra, Auli, and Rush}]{chopra_abstractive_2016}
Sumit Chopra, Michael Auli, and Alexander~M. Rush. 2016.
\newblock \href {https://doi.org/10.18653/v1/N16-1012} {Abstractive {Sentence}
  {Summarization} with {Attentive} {Recurrent} {Neural} {Networks}}.
\newblock In \emph{Proceedings of the 2016 {Conference} of the {North}
  {American} {Chapter} of the {Association} for {Computational} {Linguistics}:
  {Human} {Language} {Technologies}}, pages 93--98, San Diego, California.
  Association for Computational Linguistics.

\bibitem[{Cohan et~al.(2018)Cohan, Dernoncourt, Kim, Bui, Kim, Chang, and
  Goharian}]{cohan_discourse-aware_2018}
Arman Cohan, Franck Dernoncourt, Doo~Soon Kim, Trung Bui, Seokhwan Kim, Walter
  Chang, and Nazli Goharian. 2018.
\newblock \href {https://doi.org/10.18653/v1/N18-2097} {A {Discourse}-{Aware}
  {Attention} {Model} for {Abstractive} {Summarization} of {Long} {Documents}}.
\newblock In \emph{Proceedings of the 2018 {Conference} of the {North}
  {American} {Chapter} of the {Association} for {Computational} {Linguistics}:
  {Human} {Language} {Technologies}, {Volume} 2 ({Short} {Papers})}, pages
  615--621, New Orleans, Louisiana. Association for Computational Linguistics.

\bibitem[{Devlin et~al.(2019)Devlin, Chang, Lee, and
  Toutanova}]{devlin_bert_2019}
Jacob Devlin, Ming-Wei Chang, Kenton Lee, and Kristina Toutanova. 2019.
\newblock \href {http://arxiv.org/abs/1810.04805} {{BERT}: {Pre}-training of
  {Deep} {Bidirectional} {Transformers} for {Language} {Understanding}}.
\newblock \emph{arXiv:1810.04805 [cs]}.
\newblock ArXiv: 1810.04805.

\bibitem[{Gaskell et~al.(2020)Gaskell, Baiz, Specia, Barbaroux, and
  Topham}]{gaskell_summarization_2020}
Alexander Gaskell, Dr~Pedro Baiz, Lucia Specia, Hugo Barbaroux, and Dr~Eric
  Topham. 2020.
\newblock \href
  {https://www.imperial.ac.uk/media/imperial-college/faculty-of-engineering/computing/public/1920-pg-projects/Technical-Writer-Assistant.pdf}
  {On the {Summarization} and {Evaluation} of {Long} {Documents}}.
\newblock page~98.

\bibitem[{Gidiotis and Tsoumakas(2020)}]{gidiotis_divide-and-conquer_2020}
Alexios Gidiotis and Grigorios Tsoumakas. 2020.
\newblock \href {https://doi.org/10.1109/TASLP.2020.3037401} {A
  {Divide}-and-{Conquer} {Approach} to the {Summarization} of {Long}
  {Documents}}.
\newblock \emph{IEEE/ACM Transactions on Audio, Speech, and Language
  Processing}, 28:3029--3040.

\bibitem[{Huang et~al.(2021)Huang, Cao, Parulian, Ji, and
  Wang}]{huang_efficient_2021}
Luyang Huang, Shuyang Cao, Nikolaus Parulian, Heng Ji, and Lu~Wang. 2021.
\newblock \href {http://arxiv.org/abs/2104.02112} {Efficient {Attentions} for
  {Long} {Document} {Summarization}}.
\newblock \emph{arXiv:2104.02112 [cs]}.
\newblock ArXiv: 2104.02112.

\bibitem[{Klymenko et~al.(2020)Klymenko, Braun, and
  Matthes}]{klymenko_automatic_2020}
Oleksandra Klymenko, Daniel Braun, and Florian Matthes. 2020.
\newblock \href {https://doi.org/10.5220/0009723306480655} {Automatic {Text}
  {Summarization}: {A} {State}-of-the-{Art} {Review}:}.
\newblock In \emph{Proceedings of the 22nd {International} {Conference} on
  {Enterprise} {Information} {Systems}}, pages 648--655, Prague, Czech
  Republic. SCITEPRESS - Science and Technology Publications.

\bibitem[{Kupiec et~al.(1995)Kupiec, Pedersen, and
  Chen}]{kupiec_trainable_1995}
Julian Kupiec, Jan Pedersen, and Francine Chen. 1995.
\newblock \href {https://doi.org/10.1145/215206.215333} {A trainable document
  summarizer}.
\newblock In \emph{Proceedings of the 18th annual international {ACM} {SIGIR}
  conference on {Research} and development in information retrieval - {SIGIR}
  '95}, pages 68--73, Seattle, Washington, United States. ACM Press.

\bibitem[{Lewis et~al.(2019)Lewis, Liu, Goyal, Ghazvininejad, Mohamed, Levy,
  Stoyanov, and Zettlemoyer}]{lewis_bart_2019}
Mike Lewis, Yinhan Liu, Naman Goyal, Marjan Ghazvininejad, Abdelrahman Mohamed,
  Omer Levy, Ves Stoyanov, and Luke Zettlemoyer. 2019.
\newblock \href {http://arxiv.org/abs/1910.13461} {{BART}: {Denoising}
  {Sequence}-to-{Sequence} {Pre}-training for {Natural} {Language}
  {Generation}, {Translation}, and {Comprehension}}.
\newblock \emph{arXiv:1910.13461 [cs, stat]}.
\newblock ArXiv: 1910.13461.

\bibitem[{Nallapati et~al.(2016)Nallapati, Zhou, santos, Gulcehre, and
  Xiang}]{nallapati_abstractive_2016}
Ramesh Nallapati, Bowen Zhou, Cicero Nogueira~dos santos, Caglar Gulcehre, and
  Bing Xiang. 2016.
\newblock \href {http://arxiv.org/abs/1602.06023} {Abstractive {Text}
  {Summarization} {Using} {Sequence}-to-{Sequence} {RNNs} and {Beyond}}.
\newblock \emph{arXiv:1602.06023 [cs]}.
\newblock ArXiv: 1602.06023.

\bibitem[{Nenkova(2011)}]{nenkova_automatic_2011}
Ani Nenkova. 2011.
\newblock \href {https://doi.org/10.1561/1500000015} {Automatic
  {Summarization}}.
\newblock \emph{Foundations and Trends® in Information Retrieval},
  5(2):103--233.

\bibitem[{Radford et~al.()Radford, Narasimhan, Salimans, and
  Sutskever}]{radford_improving_nodate}
Alec Radford, Karthik Narasimhan, Tim Salimans, and Ilya Sutskever.
\newblock Improving {Language} {Understanding} by {Generative}
  {Pre}-{Training}.
\newblock page~12.

\bibitem[{Rajpurkar et~al.(2016)Rajpurkar, Zhang, Lopyrev, and
  Liang}]{rajpurkar_squad_2016}
Pranav Rajpurkar, Jian Zhang, Konstantin Lopyrev, and Percy Liang. 2016.
\newblock \href {http://arxiv.org/abs/1606.05250} {{SQuAD}: 100,000+
  {Questions} for {Machine} {Comprehension} of {Text}}.
\newblock \emph{arXiv:1606.05250 [cs]}.
\newblock ArXiv: 1606.05250.

\bibitem[{Rush et~al.(2015)Rush, Chopra, and Weston}]{rush_neural_2015}
Alexander~M. Rush, Sumit Chopra, and Jason Weston. 2015.
\newblock \href {https://doi.org/10.18653/v1/D15-1044} {A {Neural} {Attention}
  {Model} for {Abstractive} {Sentence} {Summarization}}.
\newblock In \emph{Proceedings of the 2015 {Conference} on {Empirical}
  {Methods} in {Natural} {Language} {Processing}}, pages 379--389, Lisbon,
  Portugal. Association for Computational Linguistics.

\bibitem[{Saggion and Poibeau(2013)}]{poibeau_automatic_2013}
Horacio Saggion and Thierry Poibeau. 2013.
\newblock \href {https://doi.org/10.1007/978-3-642-28569-1_1} {Automatic {Text}
  {Summarization}: {Past}, {Present} and {Future}}.
\newblock In Thierry Poibeau, Horacio Saggion, Jakub Piskorski, and Roman
  Yangarber, editors, \emph{Multi-source, {Multilingual} {Information}
  {Extraction} and {Summarization}}, pages 3--21. Springer Berlin Heidelberg,
  Berlin, Heidelberg.
\newblock Series Title: Theory and Applications of Natural Language Processing.

\bibitem[{See et~al.(2017)See, Liu, and Manning}]{see_get_2017}
Abigail See, Peter~J. Liu, and Christopher~D. Manning. 2017.
\newblock \href {https://doi.org/10.18653/v1/P17-1099} {Get {To} {The} {Point}:
  {Summarization} with {Pointer}-{Generator} {Networks}}.
\newblock In \emph{Proceedings of the 55th {Annual} {Meeting} of the
  {Association} for {Computational} {Linguistics} ({Volume} 1: {Long}
  {Papers})}, pages 1073--1083, Vancouver, Canada. Association for
  Computational Linguistics.

\bibitem[{Sharma et~al.(2019)Sharma, Li, and Wang}]{sharma_bigpatent_2019}
Eva Sharma, Chen Li, and Lu~Wang. 2019.
\newblock \href {http://arxiv.org/abs/1906.03741} {{BIGPATENT}: {A}
  {Large}-{Scale} {Dataset} for {Abstractive} and {Coherent} {Summarization}}.
\newblock \emph{arXiv:1906.03741 [cs]}.
\newblock ArXiv: 1906.03741.

\bibitem[{Sutskever et~al.(2014)Sutskever, Vinyals, and
  Le}]{sutskever_sequence_2014}
Ilya Sutskever, Oriol Vinyals, and Quoc~V. Le. 2014.
\newblock \href {http://arxiv.org/abs/1409.3215} {Sequence to {Sequence}
  {Learning} with {Neural} {Networks}}.
\newblock \emph{arXiv:1409.3215 [cs]}.
\newblock ArXiv: 1409.3215.

\bibitem[{Vaswani et~al.(2017)Vaswani, Shazeer, Parmar, Uszkoreit, Jones,
  Gomez, Kaiser, and Polosukhin}]{vaswani_attention_2017}
Ashish Vaswani, Noam Shazeer, Niki Parmar, Jakob Uszkoreit, Llion Jones,
  Aidan~N. Gomez, Lukasz Kaiser, and Illia Polosukhin. 2017.
\newblock \href {http://arxiv.org/abs/1706.03762} {Attention {Is} {All} {You}
  {Need}}.
\newblock \emph{arXiv:1706.03762 [cs]}.
\newblock ArXiv: 1706.03762.

\bibitem[{Wang et~al.(2019)Wang, Singh, Michael, Hill, Levy, and
  Bowman}]{wang_glue_2019}
Alex Wang, Amanpreet Singh, Julian Michael, Felix Hill, Omer Levy, and
  Samuel~R. Bowman. 2019.
\newblock \href {http://arxiv.org/abs/1804.07461} {{GLUE}: {A} {Multi}-{Task}
  {Benchmark} and {Analysis} {Platform} for {Natural} {Language}
  {Understanding}}.
\newblock \emph{arXiv:1804.07461 [cs]}.
\newblock ArXiv: 1804.07461.

\bibitem[{Wolf et~al.(2020)Wolf, Debut, Sanh, Chaumond, Delangue, Moi, Cistac,
  Rault, Louf, Funtowicz, Davison, Shleifer, von Platen, Ma, Jernite, Plu, Xu,
  Scao, Gugger, Drame, Lhoest, and Rush}]{wolf_huggingfaces_2020}
Thomas Wolf, Lysandre Debut, Victor Sanh, Julien Chaumond, Clement Delangue,
  Anthony Moi, Pierric Cistac, Tim Rault, Rémi Louf, Morgan Funtowicz, Joe
  Davison, Sam Shleifer, Patrick von Platen, Clara Ma, Yacine Jernite, Julien
  Plu, Canwen Xu, Teven~Le Scao, Sylvain Gugger, Mariama Drame, Quentin Lhoest,
  and Alexander~M. Rush. 2020.
\newblock \href {http://arxiv.org/abs/1910.03771} {{HuggingFace}'s
  {Transformers}: {State}-of-the-art {Natural} {Language} {Processing}}.
\newblock \emph{arXiv:1910.03771 [cs]}.
\newblock ArXiv: 1910.03771.

\bibitem[{Zaheer et~al.(2021)Zaheer, Guruganesh, Dubey, Ainslie, Alberti,
  Ontanon, Pham, Ravula, Wang, Yang, and Ahmed}]{zaheer_big_2021}
Manzil Zaheer, Guru Guruganesh, Avinava Dubey, Joshua Ainslie, Chris Alberti,
  Santiago Ontanon, Philip Pham, Anirudh Ravula, Qifan Wang, Li~Yang, and Amr
  Ahmed. 2021.
\newblock \href {http://arxiv.org/abs/2007.14062} {Big {Bird}: {Transformers}
  for {Longer} {Sequences}}.
\newblock \emph{arXiv:2007.14062 [cs, stat]}.
\newblock ArXiv: 2007.14062.

\bibitem[{Zeng et~al.(2016)Zeng, Luo, Fidler, and
  Urtasun}]{zeng_efficient_2016}
Wenyuan Zeng, Wenjie Luo, Sanja Fidler, and Raquel Urtasun. 2016.
\newblock \href {http://arxiv.org/abs/1611.03382} {Efficient {Summarization}
  with {Read}-{Again} and {Copy} {Mechanism}}.
\newblock \emph{arXiv:1611.03382 [cs]}.
\newblock ArXiv: 1611.03382.

\bibitem[{Zhang et~al.(2020)Zhang, Zhao, Saleh, and Liu}]{zhang_pegasus_2020}
Jingqing Zhang, Yao Zhao, Mohammad Saleh, and Peter~J. Liu. 2020.
\newblock \href {http://arxiv.org/abs/1912.08777} {{PEGASUS}: {Pre}-training
  with {Extracted} {Gap}-sentences for {Abstractive} {Summarization}}.
\newblock \emph{arXiv:1912.08777 [cs]}.
\newblock ArXiv: 1912.08777.

\bibitem[{Zhu et~al.(2021)Zhu, Hinthorn, Xu, Zeng, Zeng, Huang, and
  Jiang}]{zhu_enhancing_2021}
Chenguang Zhu, William Hinthorn, Ruochen Xu, Qingkai Zeng, Michael Zeng,
  Xuedong Huang, and Meng Jiang. 2021.
\newblock \href {https://doi.org/10.18653/v1/2021.naacl-main.58} {Enhancing
  {Factual} {Consistency} of {Abstractive} {Summarization}}.
\newblock In \emph{Proceedings of the 2021 {Conference} of the {North}
  {American} {Chapter} of the {Association} for {Computational} {Linguistics}:
  {Human} {Language} {Technologies}}, pages 718--733, Online. Association for
  Computational Linguistics.

\end{thebibliography}
\bibliographystyle{acl_natbib}




\end{document}